\documentclass[conference,a4paper]{IEEEtran}

\usepackage{graphicx}
\usepackage{color}
\usepackage{amsmath}
\usepackage{comment}
\usepackage{fancyhdr}
\IEEEoverridecommandlockouts

\title{\LARGE \bf
Admittance-Based Motion Planning with Vision-Guided Initialization for Robotic Manipulators in Self-Driving Laboratories
}

\author{ Shifa Sulaiman*, Tobias Jensen, Francesco Schetter, and Simon Bøgh  
\thanks{This work was supported by the Pioneer Center for Accelerating P2X Materials Discovery, CAPeX. Shifa Sulaiman, Tobias Jensen, and Simon Bøgh are with the Department of Electronics Systems, Aalborg University, Denmark. Francesco Schetter is with the Department of Information Technology and Electrical Engineering, Università degli Studi di Napoli Federico II, Napoli, Italy \*{\tt\small *ssajmech@gmail.com}}

}

\begin{document}

\maketitle
\thispagestyle{fancy}

\begin{abstract}
Self-driving laboratories (SDLs) represent highly automated research environments that leverage advanced technologies to conduct experiments and analyze data with minimal human involvement. 
SDLs often involve delicate laboratory equipment, unexpected environmental interactions, and occasional human intervention, which makes compliant and force-aware control essential for ensuring safety, adaptability, and reliability. 
This paper introduces a motion planning framework centered on admittance control to enable adaptive and compliant robotic manipulation. Unlike conventional schemes, the proposed approach employs an admittance controller directly within trajectory execution, allowing the manipulator to dynamically respond to external forces during interaction. The capability enables human operators to override or redirect the robot’s motion in real time. A vision algorithm based on structured planar pose estimation is employed to detect and localize textured planar objects through feature extraction, homography estimation, and depth fusion, thereby providing an initial target configuration for motion planning. This initialization from vision establishes the reference trajectory, while the embedded admittance controller ensures that execution of this trajectory remains safe, adaptive, and responsive to external forces or human intervention. The proposed strategy is validated using textured image detection as a proof-of-concept, and future work will extend the framework to SDLs environments employing transparent lab objects, where compliant motion planning can further enhance autonomy, safety, and human–robot collaboration.

\end{abstract}

\section{INTRODUCTION }
Self-driving laboratories (SDLs) are emerging as transformative research environments that integrate robotics, automation, and artificial intelligence to accelerate scientific discovery \cite{p1}. By minimizing human intervention in experimental workflows, SDLs enable reproducibility, scalability, and efficiency across diverse domains such as chemistry, materials science, and biology. Integrating robotic manipulators into laboratory environments significantly enhances the efficiency and precision of synthesis experiments. 

Despite recent advances, most frameworks lack mechanisms for compliant interaction, limiting their ability to adapt when external forces are applied such as when a human operator attempts to redirect the robot’s trajectory. Vision-based tracking methods have matured considerably, yet their integration with compliance-aware motion planning strategies remains underexplored in SDLs contexts. This paper addresses these gaps by presenting a motion planning framework that integrates admittance control with vision-guided object tracking. 
The contributions of the work are as follows:

\begin{itemize}
  \item \textbf{Compliance-integrated motion planning:}
  A motion planning scheme that embeds an admittance control into trajectory execution, enabling a manipulator to adaptively respond to external forces and ensuring safe human-robot collaboration.
  The framework allows a human to physically redirect the manipulator away from its goal pose and supports resuming or adjusting the trajectory after force interaction capabilities not feasible without non-compliant control.

  \item \textbf{Vision-guided proof-of-concept with SDLs relevance:}
  Validation using a structured planar pose estimation algorithm for textured object detection, with the long-term contribution focused on extending compliance-aware motion planning to SDL environments for enhanced autonomy and safety.
\end{itemize}

\section{BACKGROUND}

Robots are increasingly deployed in pharmaceutical and chemical laboratories to automate logistics and experimental workflows. A system given in \cite{ref5} combined a robotic arm, refrigerated storage, and 2D/3D vision to transport SBS-format plates and interfaced with analytical equipment, enabling intra- and inter-lab operations with reduced manual handling. 
NIST introduced a configurable mobile manipulator \cite{ref7} using coordinate registration, Kalman filtering, and optical tracking to benchmark pose repeatability in large-scale laboratory settings. Jiang \textit{et al.} \cite{ref11_new} demonstrated a dual-arm manipulator capable of precise solid dispensing at milligram scales, highlighting the potential of multi-arm systems for fine laboratory operations.

Admittance control enables compliant interaction in a rigid \cite{new,new1}  or soft \cite{new2,new3} robots, allowing manipulators to adapt to external forces while maintaining safe autonomous execution. Fukumoto \textit{et al.} \cite{adm_1} presented a mobile manipulator for door opening/closing tasks using admittance control and force decomposition, achieving reduced tip force and improved stability, though real-time dynamic control and task generalization remained challenging. Sidiropoulos \textit{et al.} \cite{adm_2} proposed a variable admittance scheme for cooperative manipulation of high-inertia objects, dynamically adjusting damping based on transmitted power. 
 Gholampour \textit{et al.} \cite{adm_3} introduced a mass-adaptive admittance framework with real-time payload estimation, compensating for sagging and improving trajectory tracking on a UR5e robot. Rhee \textit{et al.} \cite{adm_4} developed a hybrid impedance–admittance strategy that switched based on trajectory and error thresholds, ensuring stable interaction across stiff and soft environments. Tarbouriech \textit{et al.} \cite{adm_5} proposed a hierarchical quadratic programming framework for dual-arm cobots, integrating admittance control for safe bimanual manipulation, validated on the BAZAR platform. Huang \textit{et al.} \cite{adm_6} designed an admittance framework for dual-arm cooperative tasks in variable environments, using force feedback and simulation-based tuning of inertia, damping, and stiffness parameters.

 \section{Mobile Manipulator}
\begin{figure}[hbt!]
    \centering
    \includegraphics[width=0.4\textwidth]{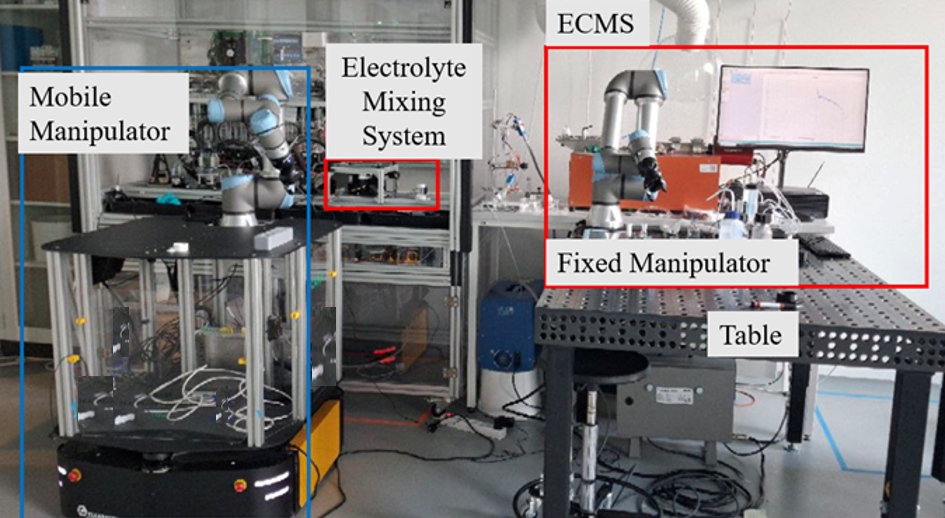}
    \caption{Lab environment with a mobile manipulator, a fixed manipulator, and other laboratory equipments}
    \label{fig1}
\end{figure}

 The experimental platform employed in this study was a mobile manipulator constructed on the Ridgeback omni‑directional base from Clearpath Robotics (shown in Fig. \ref{fig1}). The Ridgeback provides holonomic motion and high payload capacity, enabling precise maneuvering in cluttered laboratory environments. Mounted on this platform, a Universal Robots (UR5e) collaborative arm offers six degrees of freedom (DOF), high positional accuracy, and integrated force‑torque sensing, making it well suited for delicate manipulation tasks. At the end‑effector, a Robotiq Hand‑E adaptive gripper delivers versatile grasping capabilities, accommodating a wide range of laboratory objects from small vials to larger containers. 
To enable perception and workspace awareness, the mobile manipulator was equipped with two LiDAR sensors and three Intel RealSense cameras. These sensors provide rich spatial and visual data, supporting localization and environment mapping.

\section{Admittance control of the manipulator}

\begin{figure}[hbt!]
    \centering
    \includegraphics[width=0.48\textwidth]{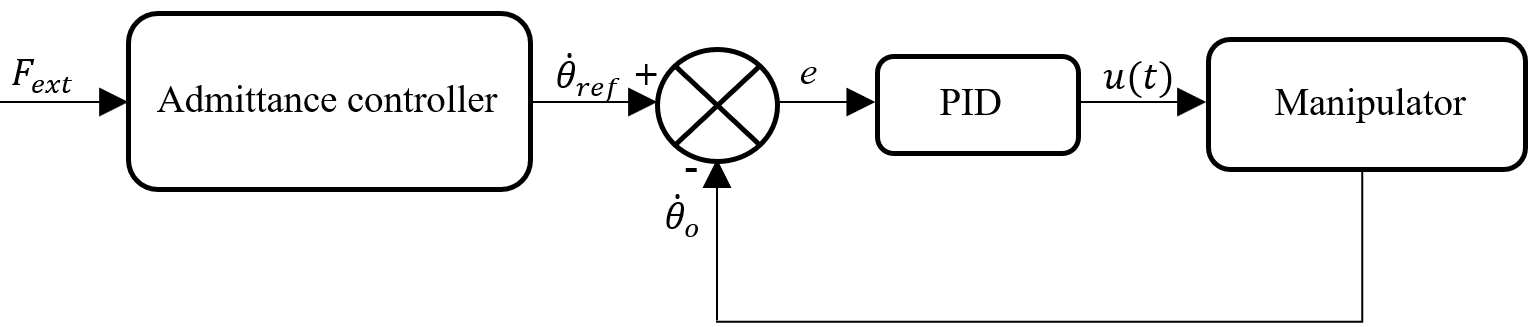}
    \caption{Admittance control scheme for the manipulator}
    \label{control_scheme}
\end{figure}

Admittance control establishes a dynamic mapping from external force input to motion output. 
The general continuous-time model is expressed as given in Eq. \eqref{eq10}
\begin{equation}
M_d \ddot{x}(t) + B_d \dot{x}(t) + K_d \big(x(t) - x_d(t)\big) = F_{\text{ext}}(t)
\label{eq10}
\end{equation}
where $M_d$, $B_d$, $K_d$, $x(t)$, $\dot x(t)$, $\ddot x(t)$, $x_d(t)$, and $F_{\text{ext}}(t)$ represent desired inertia matrix, desired damping matrix, desired stiffness matrix, end-effector position, end-effector velocity, end-effector acceleration, desired  end-effector position, and external force respectively. The Cartesian reference velocity ($\dot{x}_{\text{ref}}(t)$) is computed based on Eq. \eqref{eq11}
\begin{equation}
\dot{x}_{\text{ref}}(t) = M_d^{-1} \Big( F_{\text{ext}}(t) - B_d \dot{x}(t) - K_d (x(t) - x_d(t)) \Big)
\label{eq11}
\end{equation}
The Cartesian reference velocity is mapped to joint space velocity ($\dot{\theta}_{\text{ref}}$) using the manipulator Jacobian ($J^{\dagger}(\theta)$) as given in Eq. \eqref{eq12}
\begin{equation}
\dot{x}_{\text{ref}} = J^{\dagger}(\theta) \, \dot{\theta}_{\text{ref}}
\label{eq12}
\end{equation}
The Joint space velocity ($\dot{\theta}_{\text{ref}}$) is computed using Eq. \eqref{eq_12}
\begin{equation}
\dot{\theta}_{\text{ref}} = (J^{\dagger})^{-1}(\theta) \, \dot{x}_{\text{ref}}
\label{eq_12}
\end{equation}
Finally, the commanded trajectory ($x_{\text{cmd}}(t)$) blends the nominal plan with the admittance displacement ($x_{\text{adm}}(t)$) as given in Eq. \eqref{eq13}
\begin{equation}
x_{\text{cmd}}(t) = x_d(t) + \Delta x_{\text{adm}}(t)
\label{eq13}
\end{equation}
For stability and passivity, the parameters must satisfy Eq. \eqref{eq14}
\begin{equation}
M_d \succ 0, \quad B_d \succ 0, \quad K_d \succeq 0
\label{eq14}
\end{equation}
The proposed control scheme for a manipulator is shown in Fig. \ref{control_scheme}.
The proposed control architecture integrates admittance control with a closed-loop velocity regulation strategy to enable compliant manipulation. As illustrated in Fig.~\ref{control_scheme}, the system established a dynamic mapping from external force input ($F_{\text{ext}}(t)$) to motion output, allowing the manipulator to adapt its trajectory in response to physical interaction.
The control loop started with the admittance control scheme, which implemented the continuous-time model defined in equation~\eqref{eq10}. This model treated the manipulator as a virtual mass-spring-damper system, where the desired inertia ($M_d$), damping ($B_d$), and stiffness ($K_d$) govern the system’s responsiveness to external forces. The controller computed the Cartesian reference velocity ($\dot{x}_{\text{ref}}(t)$) using equation~\eqref{eq11}, reflecting the system’s dynamic response to the applied force.
To execute this velocity in joint space, the Cartesian velocity is mapped via the manipulator Jacobian using equation~\eqref{eq_12}, yielding the desired joint velocity $\dot{\theta}_{\text{ref}}$. This velocity served as the reference input for the low-level velocity control loop.
The summing junction compared the desired joint velocity ($\dot{\theta}_{\text{ref}}$) with the actual joint velocity ($\dot{\theta}_0$), producing a velocity error signal ($e(t)$) given in Eq. \eqref{eq_1}
\begin{equation}
    e(t) = \dot{\theta_{ref}} - \dot{\theta_{o}}
    \label{eq_1}
\end{equation}
This error is fed into a PID controller, which generated a control signal $u(t)$ to drive the manipulator as given in Eq.  \eqref{eq_2}
\begin{equation}
    u(t) = K_Pe(t)+K_I\int_o^t e(t)dt~+K_D\dot{e}(t)
    \label{eq_2}
\end{equation}
The PID controller ensured that the actual joint velocity tracked the reference velocity with minimal error, compensating for system dynamics and disturbances.
The manipulator received the control signal ($u(t)$) and executed the motion, closing the feedback loop. The actual joint velocity ($\dot{\theta}_0$) is continuously monitored and fed back to the summing junction, maintaining real-time compliance.
Finally, the commanded Cartesian trajectory ($x_{\text{cmd}}(t)$) is computed by blending the nominal trajectory $x_d(t)$ with the admittance-induced displacement ($\Delta x_{\text{adm}}(t)$), as shown in equation~\eqref{eq13}. This blending ensured that the manipulator followed its planned path while remaining responsive to external interactions.
To guarantee system stability and passivity, the admittance parameters must satisfy the conditions in equation~\eqref{eq14}, ensuring positive definiteness of inertia and damping matrices and non-negativity of stiffness.

\section{Vision-Based Grasping Framework}

The vision-based algorithm \cite{r1} facilitated robust object manipulation by coupling visual perception with adaptive grasp planning. The framework was organized into two principal modules:

\begin{itemize}
  \item Object detection and planar pose estimation
  \item Adaptive grasp generation and refinement
\end{itemize}

\subsection{Object Detection and Pose Estimation}

Reliable robotic manipulation requires precise estimation of object pose. The proposed method was structured into four sequential stages:

\begin{enumerate}
  \item Extraction of visual features and descriptor matching
  \item Homography computation with perspective transformation
  \item Construction of the object coordinate frame using directional vectors
  \item Pose refinement through depth sensing
\end{enumerate}

Planar objects were detected using the Scale-Invariant Feature Transform (SIFT), which identified distinctive keypoints and descriptors to enable robust recognition and matching. Descriptor correspondences were established using Fast Library for Approximate Nearest Neighbors (FLANN) for floating-point descriptors or Hamming distance for binary descriptors, ensuring accurate alignment between the input image and a reference template.
From the matched keypoints, a homography matrix $\mathbf{H}$ was estimated to model the planar transformation as given in Eq. \eqref{eq1}
\begin{equation}
\begin{bmatrix}
a \\ 
b \\ 
c
\end{bmatrix}
=
\mathbf{H}
\begin{bmatrix}
x \\ 
y \\ 
1
\end{bmatrix}
\label{eq1}
\end{equation}
The transformed coordinates $(x', y')$ were then computed as given in Eq. \eqref{eq2}
\begin{equation}
x' = \frac{a}{c}, \quad y' = \frac{b}{c}
\label{eq2}
\end{equation}
Random Sample Consensus (RANSAC) was employed to eliminate outliers, thereby ensuring robust homography estimation.
To establish the local coordinate frame of the object, three reference points were defined in Eq.  \eqref{eq3}
\begin{equation}
P_c = (w/2, h/2), \quad P_x = (w, h/2), \quad P_y = (w/2, 0)
\label{eq3}
\end{equation}
where $w$ and $h$ denote the object’s width and height. These points were projected into 3D using RGB-D data from the RealSense camera, yielding directional vectors given in Eq. \eqref{eq4}
\begin{equation}
\vec{i} = \frac{\vec{x}}{\|\vec{x}\|}, \quad \vec{j} = \frac{\vec{y}}{\|\vec{y}\|}, \quad \vec{k} = \frac{\vec{x} \times \vec{y}}{\|\vec{x} \times \vec{y}\|}
\label{eq4}
\end{equation}
The orthonormal basis $(\vec{i}, \vec{j}, \vec{k})$ defined the orientation of the object in space.
The rotation matrix $R$ was constructed from these vectors as given in Eq. \eqref{eq5}
\begin{equation}
R =
\begin{bmatrix}
i_x & j_x & k_x \\
i_y & j_y & k_y \\
i_z & j_z & k_z
\end{bmatrix}
\label{eq5}
\end{equation}
Euler angles $(\phi, \theta, \psi)$ were then derived as given in Eq. \eqref{eq6}
\begin{equation}
\theta = \tan^{-1}(j_z), \quad \phi = \sin^{-1}(-i_z), \quad \psi = \tan^{-1}\left(\frac{i_y}{i_x}\right)
\label{eq6}
\end{equation}
These angles represented the object’s orientation relative to the global reference frame and were subsequently used for grasp planning.

\section{Results and Discussion}
Simulation studies and experimental validations were conducted to assess the effectiveness of the proposed controller during a grasping task. The outputs generated by the vision module were subsequently employed to guide and manipulate the motions of the mobile manipulator.

\subsection{Simulation Study}

The admittance control parameters were determined through a combination of theoretical modeling and empirical tuning. The desired inertia matrix was set to $M_{d} = 5 \,\text{kg}$, reflecting the effective mass of the manipulator’s end-effector and gripper assembly. The damping coefficient was chosen as $B_{d} = 50 \,\text{Ns/m}$, derived from iterative stability analysis to ensure smooth velocity regulation without excessive oscillations. The stiffness parameter was fixed at $K_{d} = 100 \,\text{N/m}$, based on workspace compliance requirements and the need to maintain positional accuracy during contact. These values were obtained by benchmarking the manipulator’s dynamic response under small perturbations and verifying passivity conditions, thereby ensuring that the system remained stable while preserving compliant behavior. 
The proposed method was implemented within the ROS framework on an Ubuntu $20.04$ platform. The experiments were conducted using a $3.0$ GHz Intel Core i7-7400 CPU equipped with 16GB of RAM. A book with textured front cover was used as a textured object. The trajectory was determined using the  Rapidly-exploring Random Tree star (RRT*) algorithm, while the inverse kinematic solutions were obtained through the Damped Least Square (DLS) approach.

\begin{figure}[hbt!]
    \centering
    \includegraphics[width=0.2\textwidth]{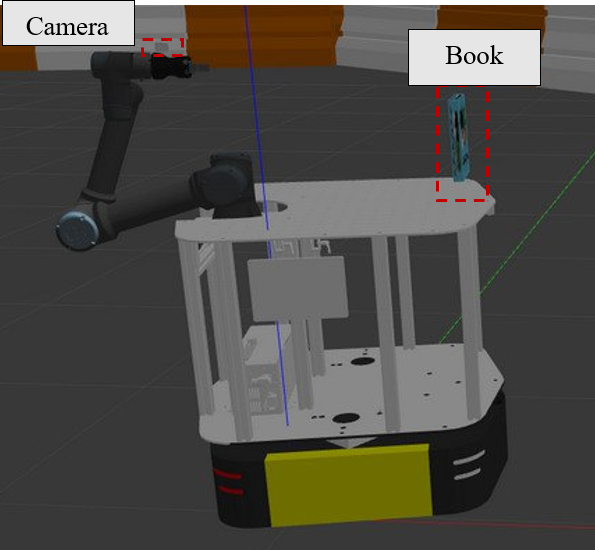}
    \caption{Simulation setup for vision-guided grasping}
    \label{sim_setup}
\end{figure}
\begin{figure}[hbt!]
    \centering
    \includegraphics[width=0.25\textwidth]{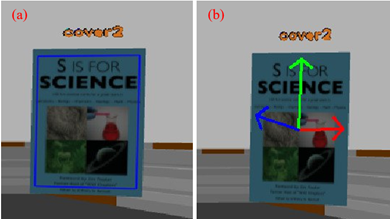}
    \caption{Output of vision algorithm (a)Boundary box (b)Axes of frames}
    \label{vision}
\end{figure}
\begin{figure}[hbt!]
    \centering
    \includegraphics[width=0.4\textwidth]{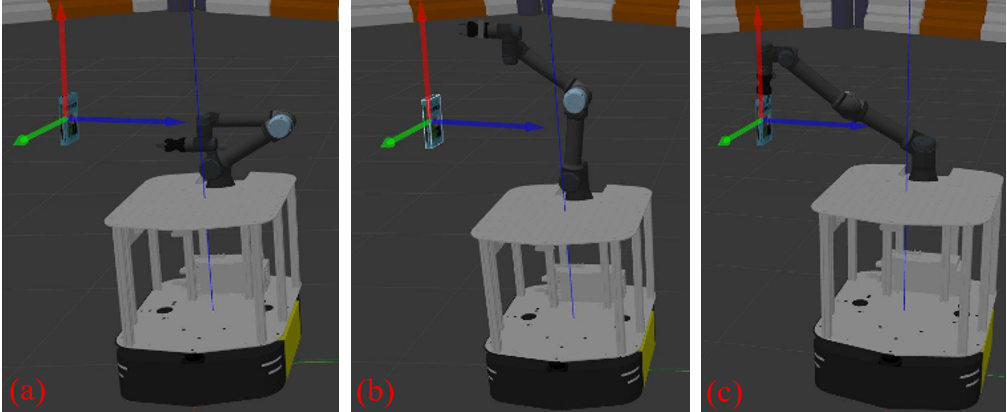}
    \caption{Motion of manipulator to grasp a book during simulation}
    \label{motion}
\end{figure}
\begin{figure}[hbt!]
    \centering
    \includegraphics[width=0.45\textwidth]{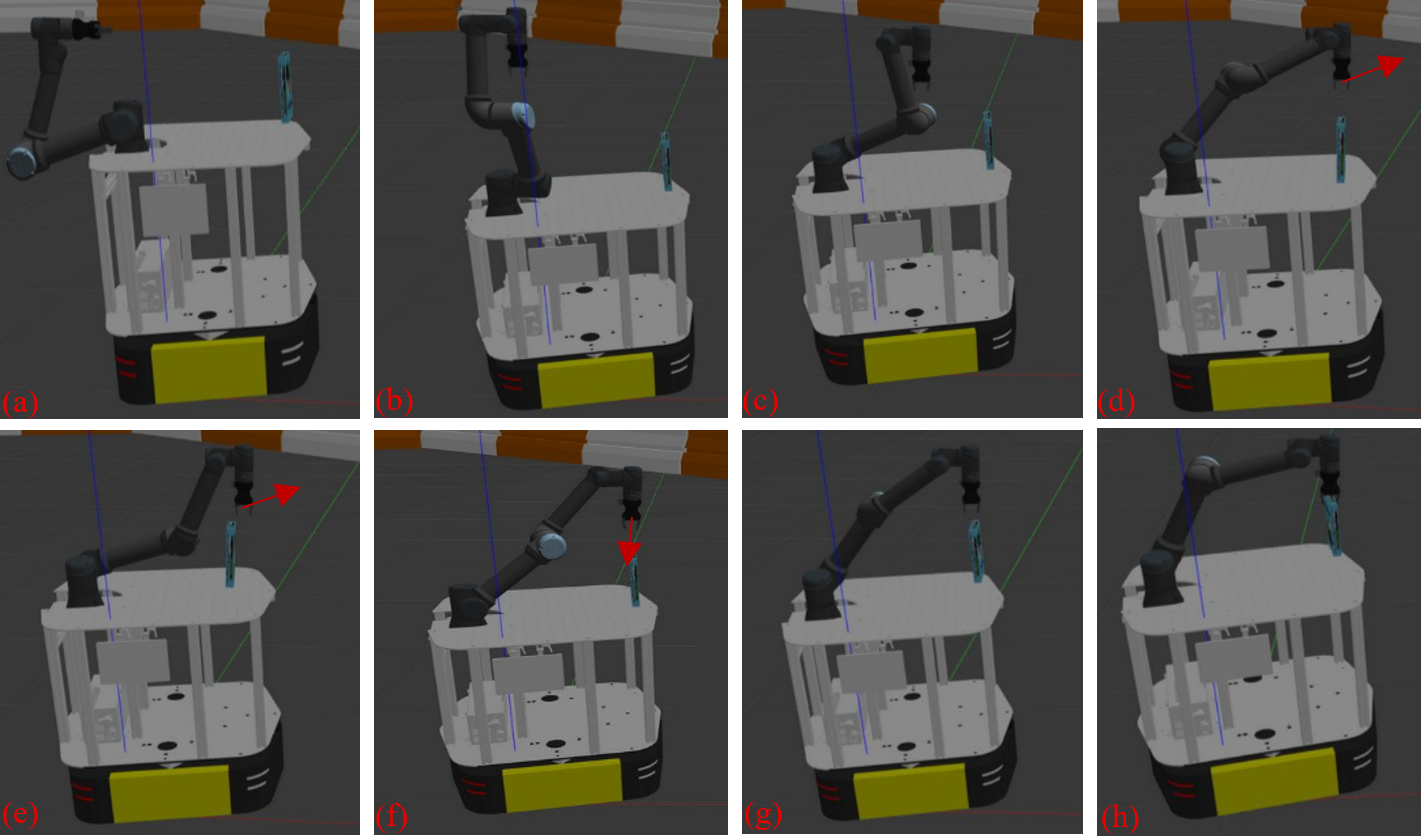}
    \caption{Manipulator motion under admittance control with external force applied}
    \label{motion1}
\end{figure}
\begin{figure}[hbt!]
    \centering
    \includegraphics[width=0.45\textwidth]{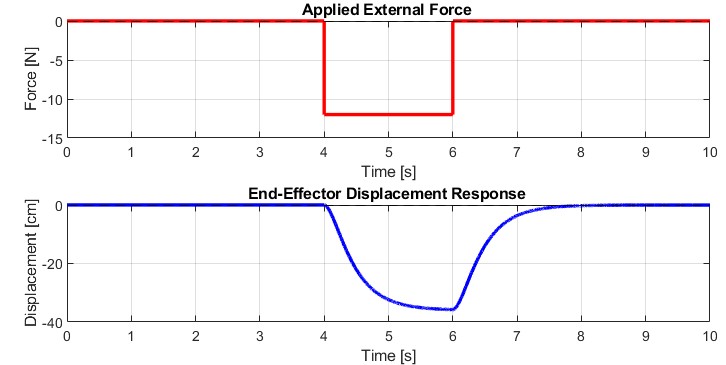}
    \caption{Simulation results (a)Force (b)Displacement }
    \label{graphs}
\end{figure}

A RealSense camera mounted on the manipulator’s end effector (shown in Fig. \ref{sim_setup}) was employed to detect the pose of the textured front cover of a book. The RRT* motion planning algorithm then utilized this pose information to grasp the book using the vision algorithm. The outputs of the vision algorithm are shown in Figs. \ref{vision} (a) and (b), where Fig.\ref{vision} (a) illustrates the detected bounding box around the object, and Fig. \ref{vision} (b) depicts the estimated pose of book cover.
The motion of the manipulator during the grasping task is illustrated in Fig.~\ref{motion}. After receiving pose of the book, the manipulator moves forward to grasp the book as shown in Figs. \ref{motion} (a) - (c). In this scenario, no external force was applied, thereby allowing the admittance control to remain inactive throughout the process.
The second case involved the application of an external force, as illustrated in Fig.~\ref{motion1}.
 During the grasping attempt, a force of 12 N was applied in the 
-y direction for a time duration of 2 seconds, which induced motion of the manipulator consistent with the applied force, as shown in the Figs. \ref{motion1} (d) and (e). Following the force application, the manipulator returned to its planned trajectory (Fig.~\ref{motion1} (f)) and successfully grasped the object, as depicted in Figs.~\ref{motion1} (g) - (h). 
Variations in force and displacement with respect to its desired values are shown in Figs. \ref{graphs} (a) and (b) respectively. These results validated the theoretical admittance model (mass–spring–damper analogy) by showing that a smooth motion of manipulator reacting in a proportional and predictable way to external forces.
Instead of resisting or failing, the robot yields to human intervention and then smoothly returns to its planned trajectory. The results confirmed that the embedded admittance controller enabled compliant interaction, which is crucial for SDLs.
Performance metric of the vision algorithm is given in Table \ref{tab1}. Ground-truth poses were established via ROS Gazebo reference frames, and all estimated poses were expressed in a common coordinate frame to enable direct comparison. occlusion occurred due to end-effector constraints. The grasping success rate was measured at 95.3\%, demonstrating robust manipulation capabilities even under partial occlusions and pose drift. Pose estimation error remained within $\pm 0.43$~cm, reflecting high spatial precision suitable for fine-grained grasping tasks. The detection pipeline operated with an average latency of 74~ms per frame, supporting real-time responsiveness at 12~FPS. Detection precision and recall were recorded at 96.5\% and 98.7\%, respectively, validating the system's effectiveness in identifying and localizing target objects under variable lighting and background clutter. 
Overall, these results confirmed the framework’s suitability for manipulation tasks in semi-structured environments, with strong generalization across object types and camera viewpoints. 
\begin{table}[htbp]
\centering
\caption{Performance Metrics for Object Tracking and Grasping}
\begin{tabular}{|l|c|c|}
\hline
\textbf{Metric} & \textbf{Simulation Value} & \textbf{Experimental Value} 
\\
\hline
Grasping Success Rate (\%) & 95.3 & 93.7 
\\
\hline
Pose Estimation Error (cm) & $\pm$0.43  & $\pm$0.61  
\\
\hline
Detection Latency (ms) & 74  & 71  
\\
\hline
Detection Precision (\% ) & 96.5 & 93.1 
\\
\hline
Detection Recall (\%) & 98.7 & 97.5 
\\
\hline
Runtime Performance (FPS) & $\sim$12  & $\sim$13  
\\
\hline
\end{tabular}
\label{tab1}
\end{table}
\subsection{Experimental validation}
\begin{figure}[hbt!]
    \centering
    \includegraphics[width=0.3\textwidth]{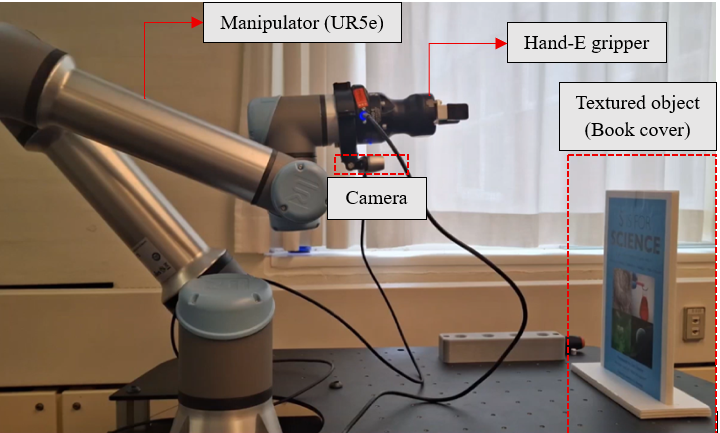}
    \caption{Experimental setup for vision-guided compliant grasping}
    \label{Experimentation setup}
\end{figure}

\begin{figure}[hbt!]
    \centering
    \includegraphics[width=0.45\textwidth]{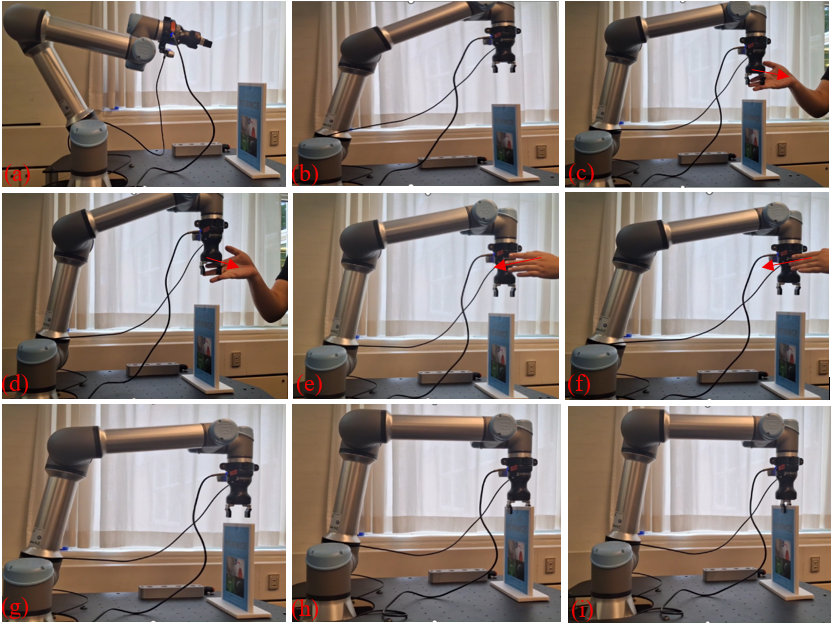}
    \caption{Manipulator motion during experimental grasping trials}
    \label{exp_motion}
\end{figure}

\begin{figure}[hbt!]
    \centering
    \includegraphics[width=0.45\textwidth]{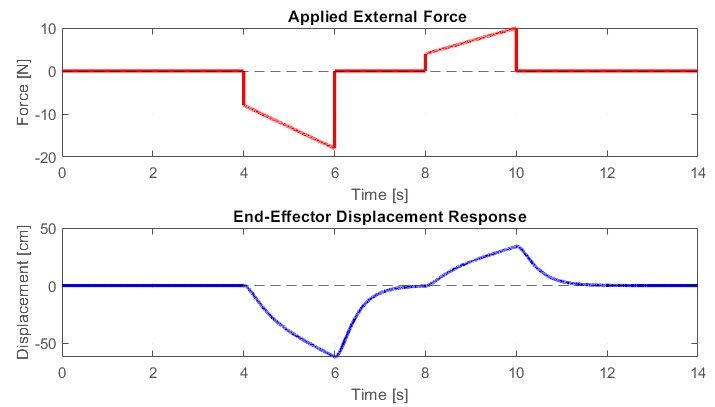}
    \caption{Experimentation results (a)Force (b)Displacement }
    \label{graphs1}
\end{figure}

Fig.~\ref{Experimentation setup} depicts the experimental setup designed to evaluate the robotic system’s object interaction capabilities. The configuration consisted of a robotic manipulator mounted on a mobile base and positioned in front of a textured book cover. This arrangement replicated the simulation conditions, enabling a direct comparison between predicted and observed manipulator performance.
Fig. \ref{exp_motion} (a) – (i) demonstrate the manipulator’s ability to track the textured object across multiple spatial configurations. Figs.  \ref{exp_motion} (a) and (b) show the motion of manipulator to grasp the book after receiving the pose of book from the vision algorithm. We applied a force upto 10 N in +y direction and 18 N in -y direction. Upon physical interaction, the robot yielded to the applied force, deviated from its nominal trajectory, and subsequently resumed its planned motion once the force was removed (refer attached video). The sequence of subfigures \ref{exp_motion} (c) – (f) show the robot arm dynamically adjusting its position as the application of force changes. Figs. \ref{exp_motion} (g) – (i) illustrate the execution of a grasping maneuver, culminating in a successful grip of the book. This sequence emphasized the effectiveness of the grasp planning module, which utilized pose data and trajectory optimization to achieve stable contact. The successful execution of grasping action by complying with the external force confirmed the capability of the proposed admittance controller along with pose estimation strategy. The consistency between simulated grasp strategies and experimental outcomes further reinforced the reliability of the system’s control architecture.The experimental results further showcased the effectiveness of the proposed admittance-based control framework. As shown in Figs. \ref{graphs1}, the manipulator demonstrated compliant behavior when subjected to external forces during grasping tasks. This smooth behavior confirmed the system’s ability to adapt in real-world settings, maintaining both safety and task fidelity. The consistency between simulated and experimental responses validated the robustness of the admittance controller and its suitability where human–robot collaboration and adaptive manipulation are essential.

As summarized in Table~\ref{tab1}, the proposed framework achieved a grasping success rate of 93.7\%, confirming robust manipulation performance. The pose estimation error was maintained within $\pm 0.61$~cm, demonstrating high spatial precision suitable for fine-grained grasping tasks. The detection pipeline operated with an average latency of 71~ms per frame, corresponding to real-time responsiveness at approximately 13~FPS. Furthermore, detection precision and recall were measured at 93.1\% and 97.5\%, respectively, validating the effectiveness of the system in reliably identifying and localizing objects across diverse lighting conditions and background clutter.

\subsection{Extended Quantitative Evaluation of Admittance Controller}

To complement the vision-based metrics, additional quantitative measures were introduced to evaluate the admittance controller’s performance in both simulation and experimental trials as shown in Table \ref{tab2}. In simulation, the force tracking error remained within 0.07~N, confirming close adherence to the theoretical mass-spring-damper model. Experimental trials yielded a slightly higher error of 0.13~N, reflecting sensor noise and calibration uncertainties. 
The maximum deviation of manipulator due to the application of external force is obtained as 0.31~cm in simulation and 1.02~cm in experiments, with recovery times of 1.11~s and 2.08~s, respectively. These results highlight the controller’s ability to balance compliance with trajectory fidelity across both domains. 
Dynamic response analysis revealed a settling time of 1.11~s in simulation and 2.08~s experimentally, with damping ratios of 0.62 and 0.58, respectively. The slightly lower damping ratio in experiments indicates minor oscillatory tendencies due to real-world mechanical tolerances. The energy dissipation during interaction was quantified by integrating force-displacement profiles. For a 10~N perturbation, the dissipated mechanical energy was 0.21~J in simulation and 0.25~J experimentally, confirming the controller’s ability to absorb external energy without destabilization. 
Motion smoothness was assessed through velocity overshoot and jerk analysis. Overshoot averaged 4.80\% in simulation and 6.30\% experimentally, while peak jerk magnitudes were 0.42~m/s$^{3}$ and 0.59~m/s$^{3}$, respectively. These values confirm that the manipulator’s motion remained continuous and human-compatible. 
Finally, human effort reduction was quantified by comparing operator force requirements with and without admittance control. In simulation, effort was reduced by 38\%, while experimental trials achieved a reduction of 34\%, validating ergonomic benefits in real-world collaborative scenarios.

\begin{table}[htbp]
\centering
\caption{Comparative Metrics for Admittance Controller Performance}
\begin{tabular}{|l|c|c|}
\hline
\textbf{Metric} & \textbf{Simulation} & \textbf{Experiment} 
\\ \hline
Force Tracking Error (RMSE) (N) & 0.07 & 0.13 
\\ \hline
Maximum Trajectory Deviation (cm) & 0.31 & 1.02 
\\ \hline
Recovery Time (s) & 0.80 & 1.21 
\\ \hline
Settling Time (s) & 1.11 & 2.08 
\\ \hline
Damping Ratio & 0.62 & 0.58 
\\ \hline
Energy Dissipation (J) & 0.21 & 0.25 
\\ \hline
Velocity Overshoot (\%) & 4.80 & 6.30 
\\ \hline
Peak Jerk (m/s$^{3}$) & 0.42 & 0.59 
\\ \hline
Human Effort Reduction (\%) & 38 & 34 
\\
\hline

\end{tabular}
\label{tab2}
\end{table}
This comparative analysis showed strong consistency between simulation and experimental results, with only minor deviations attributable to real-world uncertainties such as sensor noise, mechanical tolerances, and environmental variability. Together, these metrics provide a multi-dimensional validation of the admittance controller’s robustness, safety, and ergonomic benefits in SDL contexts.

\section{Conclusion}
In this study, the proposed motion planning strategy is validated through a proof‑of‑concept demonstration employing an admittance controller with a textured image detection. 
Simulation and experimental results confirmed the effectiveness of the framework. The manipulator demonstrated smooth and proportional responses to applied forces, validating the admittance model’s mass–spring–damper analogy. In experimental trials, the manipulator responded to forces up to 18 N without instability, confirming real-time compliance and robustness. Simulation results related to vision method provided idealized conditions, leading to slightly higher success rates and lower pose estimation error. In simulation, cameras and depth sensors are modeled without noise, distortion, or calibration drift. These led to more accurate feature extraction and pose estimation. However, the performance is acceptable due to the robustness of the vision pipeline, which maintained high detection precision and recall even under real-world conditions.
Future work will focus on extending the framework to handle transparent and deformable laboratory objects, integrating continuous visual feedback for dynamic pose tracking, and scaling the system to multi-arm or multi-agent configurations. These enhancements aim to further improve autonomy, safety, and task generalization in SDLs, ultimately accelerating scientific workflows through intelligent and compliant robotic manipulation.

\end{document}